\documentclass{article}

\usepackage[preprint]{neurips_2024}

\makeatletter
\let\NeurIPS@addfooter\relax
\makeatother

\usepackage[utf8]{inputenc} 
\usepackage[T1]{fontenc}    
\usepackage{hyperref}       
\usepackage{url}            
\usepackage{booktabs}       
\usepackage{amsfonts}       
\usepackage{nicefrac}       
\usepackage{microtype}      
\usepackage{xcolor}         
\usepackage{graphicx} 
\usepackage{diagbox} 
\usepackage{svg}

\setcitestyle{round}

\title{Role-Playing Agents Driven by Large Language Models: Current Status, Challenges, and Future Trends}

\author{%
  Ye Wang$^{1}$$^{2}$\quad Jiaxing Chen$^{1}$$^{2}$\quad Hongjiang Xiao$^{1}$\thanks{Corresponding author}\\
  $^{1}$State Key Laboratory of Media Convergence and Communication, Communication University of China\\
  $^{2}$Neuroscience and Intelligent Media Institute, Communication University of China\\
  \texttt{yewang@cuc.edu.cn, xicassie6@gmail.com, xiaohj@cuc.edu.cn} 
}

\begin{document}

\maketitle

\begin{abstract}
  In recent years, with the rapid advancement of large language models (LLMs), role-playing language agents (RPLAs) have emerged as a prominent research focus at the intersection of natural language processing (NLP) and human-computer interaction. This paper systematically reviews the current development and key technologies of RPLAs, delineating the technological evolution from early rule-based template paradigms, through the language style imitation stage, to the cognitive simulation stage centered on personality modeling and memory mechanisms. It summarizes the critical technical pathways supporting high-quality role-playing, including psychological scale-driven character modeling, memory-augmented prompting mechanisms, and motivation-situation-based behavioral decision control. At the data level, the paper further analyzes the methods and challenges of constructing role-specific corpora, focusing on data sources, copyright constraints, and structured annotation processes. In terms of evaluation, it collates multi-dimensional assessment frameworks and benchmark datasets covering role knowledge, personality fidelity, value alignment, and interactive hallucination, while commenting on the advantages and disadvantages of methods such as human evaluation, reward models, and LLM-based scoring. Finally, the paper outlines future development directions of role-playing agents, including personality evolution modeling, multi-agent collaborative narrative, multimodal immersive interaction, and integration with cognitive neuroscience, aiming to provide a systematic perspective and methodological insights for subsequent research.
\end{abstract}

\section{Introduction}
With the continuous development of large language models (LLMs) such as GPT-4, Claude, and LLaMA, the design and application of agents have entered a highly anthropomorphic phase. Compared to traditional dialogue systems, modern language models not only can understand complex instructions and generate coherent text but also demonstrate reasoning capabilities across knowledge domains. This has further spurred the emergence of a rapidly growing research direction: role-playing language agents (RPLAs).

A role-playing agent is an intelligent agent capable of simulating specific character traits, linguistic styles, and even decision-making preferences. Unlike conventional chatbots that merely provide question-answering or task execution, RPLAs emphasize the consistency of personality, the rationality of behavior, and the immersion of dialogue. They can be protagonists from works of art, historical figures from ancient times, or even user-created virtual characters. This anthropomorphic capability makes RPLAs a key technology in multiple fields, including non-player characters (NPCs) in games, virtual anchors, digital humans, educational tutoring, and psychological companionship.

In recent years, related research has experienced explosive growth. For example, Zhou et al. \citeyearpar{zhou_characterglm_2024} constructed a LLM framework supporting customized social personalities, promoting the development of role-playing systems in Chinese contexts through the integration of social behaviors and character profiles; Ran et al. \citeyearpar{ran_capturing_2024} attempted to extract character traits from psychological scales to guide RPLAs in learning character inner worlds, marking a shift in modeling paradigms from superficial linguistic imitation to deep psychological construction; Xu et al. \citeyearpar{xu_character_2024} further proposed a personality-driven decision-making perspective and constructed LIFECHOICE, a large-scale dataset of character decisions in novels, exploring whether RPLAs can simulate consistent character behavioral choices.

The rise of RPLAs as a research focus is driven by both technological and social factors. Technologically, LLMs' modeling capabilities have made it feasible to imitate character linguistic styles and knowledge backgrounds; socially, there is a growing human demand for virtual personalities.Data from a report released by Business of Apps in May 2025 \citep{curry_characterai_2025} shows that Character.AI has become one of the world's most popular chatbot services, with over 25 million monthly active users and more than 18 million chatbot characters, widely used for companionship, communication, and emotional support, particularly among adolescents and the elderly living alone. In an era where AI is increasingly entering people's hearts, building a trustworthy and empathetic role agent has become a common pursuit of both industry and academia.

However, achieving high-quality role-playing simulation still faces numerous challenges. Firstly, there is a lack of a unified paradigm for constructing consistent and sustainable character settings. During long conversations, character personalities are prone to drift or even collapse. Secondly, enabling models to make behavioral decisions consistent with character motivations in specific scenarios involves capabilities such as emotional understanding and causal reasoning. Furthermore, existing evaluation metrics primarily focus on linguistic fluency, lacking comprehensive quantitative assessments of personality fidelity and behavioral logic. Psychological scales, character portraits, and subjective evaluations are emerging as new directions for RPLA assessment \citep{wang_incharacter_2024,tu_charactereval_2024}.

This paper aims to systematically collate the research status, key technologies, data construction, evaluation frameworks, and application prospects of RPLAs. It focuses on exploring the evolution of role modeling methods from linguistic style imitation to personality cognitive simulation, as well as the role of behavioral control and memory mechanisms in maintaining character consistency. Simultaneously, it pays attention to how diverse data and benchmarks support training and evaluation. Through the analysis of representative works in recent years, this paper seeks to provide a systematic perspective and research insights for the further development of this field.

\section{Evolution of Research Paradigms}
The development of RPLAs reflects the continuous advancement of NLP technology toward personality construction and interaction realism. From early rule-based template systems to modern agents built on LLMs with consistent personalities and behavioral logic, the technological evolution of RPLAs can be roughly divided into three stages: 1) rule and template-based role simulation; 2) style imitation relying on pre-trained language models; 3) cognitive simulation centered on personality-driven and decision-consistent modeling.

\subsection{Early Paradigm: Exploration of Structured Dialogue Generation Driven by Character Relationships}
Before LLMs were widely applied in role-playing scenarios, some studies had begun to move beyond simple templates and question-answer matching, exploring more structured and context-aware role modeling methods. Among them, the character-driven dialogue story continuation task proposed by Si et al. \citeyearpar{si_telling_2021} represents an important early exploration of RPLAs. This work first introduced character relationships as a core modeling element, developing a generation mechanism for multi-character dialogue narratives. Its primary goal was to predict the next character's utterance based on the relationships and personalities among characters in multi-party dialogue scenarios, thereby advancing the story.

Based on the Transformer encoder \citep{vaswani_attention_2023}, the researchers constructed a multi-task learning model with a dual-encoder structure, integrating both story continuation and character prediction tasks. The study expanded on the existing multi-party, multi-turn, narrative dialogue dataset CRD3 (Critical Role Dungeons \& Dragons Dataset), using clustering methods to model character relationships (classified as friendly, hostile, or neutral) and automatically inferring character personalities through historical utterances and interaction behaviors. This information was embedded into the context to help the model understand the relationships and tones among characters. Under the multi-task framework, the two subtasks shared parameters and were jointly optimized, significantly improving character consistency and story coherence in multi-party dialogues.

Although the system still primarily relied on a retrieval-based structure and ignored the dynamic evolution of character emotions as the plot progressed—limiting the model's creative space—its character-driven and relationship modeling approach broke through the limitations of traditional question-answer matching systems, such as the lack of personality and short-sighted context awareness. It was the first attempt to incorporate dynamic relationships among characters into the dialogue generation process, marking a significant paradigmatic shift. This work laid the conceptual foundation and provided data resources for subsequent character-centric language agent research, promoting the development of role-playing systems from linguistic imitation to behavioral modeling.

\subsection{Turning Point: LLMs Driving a Leap in Role Simulation}
The emergence of LLMs has revolutionized the paradigm of dialogue systems. Autoregressive LLMs represented by ChatGPT can learn natural language generation patterns from massive text corpora, possessing certain capabilities of style transfer and context maintenance, making natural language role simulation possible for the first time.

Especially after the advent of ChatGPT, users can instruct the model to play a role through simple prompts, such as "You are now a British nobleman; please answer the following questions elegantly." While such role prompts based on prompt engineering are superficially effective, the stability of personality and consistency of behavior remain difficult to guarantee. In multi-turn dialogues, scenario changes, or decision-making simulations, characters are prone to breaking character, revealing that linguistic style prompts alone are insufficient to support high-quality role-playing.

\subsection{Role Cognitive Modeling: Rise of Next-Generation RPLA Systems}
To overcome the limitations of style imitation, a series of RPLA frameworks focusing on personality construction, behavioral consistency, and long-term memory simulation have emerged in recent years. Representative works of this stage can be categorized in chronological order as follows: Character-LLM \citep{shao_character-llm_2023} was one of the first LLMs to incorporate role-playing capabilities into training objectives. It constructed a training corpus containing multi-character dialogues and introduced personality instruction learning, significantly improving the model's coherence in multi-role scenarios. This work emphasized the importance of role context modeling and paved the way for subsequent personality-oriented training methods. ChatHaruhi \citep{li_chatharuhi_2023} went further, proposing a memory-based control mechanism specifically for anime role-playing. By extracting character memory chunks from scripts, it controlled generation styles and behavioral decisions. Over 50,000+ rounds of Chinese-English bilingual dialogue data it constructed injected ACG (Anime/Comic/Game) cultural characteristics into the field of role simulation, demonstrating the potential of combining script-driven memory and context prompts. Ditto \citep{chen_large_2023} was a rather philosophical attempt: it argued that LLMs are inherently superpositions of personalities, and their existing potential can be activated through self-supervised alignment. The authors constructed a large-scale character self-questioning dataset containing 4,000 characters, enabling the training of RPLAs with strong robustness and personality retention without external annotations. This method represents a lightweight supervised approach that abandons personality labeling, suitable for large-scale role generalization. The CoSER project \citep{wang_coser_2025} is one of the largest and most structurally complete RPLA construction projects to date. It not only established a multimodal character dataset containing over 17,700 novel characters with rich contexts and psychological motivations but also proposed the Given-Circumstance acting paradigm, which allows the model to generate behaviors under preset plots and motivations—similar to an actor's role immersion. This work achieved a leap from performative language models to motivation-driven behavioral agents.

\subsection{Comparison of Mainstream Models: The Tug-of-War Between Closed-Source and Open-Source Models}
In terms of foundation models for building RPLAs, current mainstream approaches fall into two categories: 1) Closed-source models (e.g., GPT-4, Claude) possess strong context understanding and style imitation capabilities, performing excellently with prompt engineering support. They are suitable for few-shot/zero-shot role simulation but suffer from poor controllability and non-debuggability; 2) Open-source models (e.g., LLaMA, ChatGLM, Baichuan) can be adapted to specific roles through supervised fine-tuning or parameter-efficient training (e.g., LoRA) but require substantial role-specific data support, and generation quality is highly sensitive to data distribution. Empirical comparisons show that in Chinese contexts, CharacterGLM \citep{zhou_characterglm_2024} and the ChatGLM series are more adept at handling socio-cultural roles, while CoSER-70B \citep{wang_coser_2025} achieved a 93.47\% accuracy rate on LIFECHOICE in multi-character English novels, even outperforming GPT-4o in certain dimensions—demonstrating the great potential of open RPLA systems.

In summary, from the perspective of technological paradigm evolution, RPLAs have undergone a three-stage leap: from linguistic templates to style mimicry, and then to personality construction and behavioral simulation. Their core is no longer imitating how a character speaks but enabling the character to possess its own behavioral logic, emotional responses, and cognitive motivations. A truly high-quality RPLA system should simultaneously possess three core capabilities: linguistic consistency, behavioral rationality, and memory persistence. Future breakthroughs may lie in cross-modal character models (integrating images, sounds, and actions) and self-forming and evolving personality construction in long-term interactions. RPLAs should not merely imitate others but become new cognitive agents with "self-awareness".

\section{Core Technologies of Role-Playing Modeling}
The core capabilities of RPLAs lie not only in the fluency and diversity of language generation but also in the sustained shaping of character personality, the effective utilization of character memory, and the consistent decision-making of character behavior. To realize a credible, coherent, and motivation-driven role agent, researchers have conducted various modeling attempts from three directions: psychological modeling, memory structure design, and behavioral reasoning.

\subsection{Character Setting and Personality Modeling}
Traditional character construction mostly relies on external descriptions or script texts, such as character biographies and dialogue corpora. While this approach can shape character styles at the linguistic level, it often overlooks the modeling of personality motivations, cognitive tendencies, and behavioral preferences. To address this gap, recent works have begun to introduce personality theories from psychology, particularly the Big Five personality traits and MBTI scales.

For example, Wang et al. \citeyearpar{wang_incharacter_2024} designed a psychological test-based question-answering system with the goal of evaluating personality consistency, analyzing whether RPLAs stably exhibit certain personality dimensions (e.g., openness, conscientiousness) in multi-turn dialogues, and proposing "personality fidelity" as a new evaluation dimension. Ran et al. \citeyearpar{ran_capturing_2024} put forward the concept of personality-indicative data, using psychological scale test questions as dialogue prompts to guide LLMs in generating linguistically responsive content with personality traits, thereby training agents with personality fidelity. Their research found that features such as vocabulary preferences, sentence length, and emotional intensity in model-generated content are significantly correlated with target personality types, providing empirical support for the mapping between linguistic behavior and psychological scales.

Furthermore, PsyMem \citep{cheng_psymem_2025} proposed a more refined psychological consistency modeling framework. Breaking through the static and coarse-grained limitations of traditional personality scale inputs, it constructed a character profiling system containing 26 quantitative psychological indicators. These indicators combine latent psychological attributes (e.g., Big Five personality and values) with explicit behavioral patterns (e.g., social style, leadership tendency, and conflict resolution strategies), comprehensively depicting the character's intrinsic motivations and external behaviors. More importantly, PsyMem introduced a memory alignment training mechanism, using knowledge graphs to cognitively anchor the character's historical behaviors with current responses, ensuring that character generation is not only regulated by psychological traits but also dynamically responds to known backgrounds and memory information. This integration of modern psychological frameworks and neural mechanisms significantly improves the consistency and long-term stability of character behaviors, providing a new paradigm for credible personality simulation.

In terms of personality modeling methods, existing works can be roughly divided into two categories: supervised modeling, which relies on explicitly labeled personality tags or assessment questionnaires; and self-supervised modeling, which automatically extracts character style signals from large-scale corpora and performs alignment training. Supervised methods, such as Ran et al. \citeyearpar{ran_capturing_2024} using MBTI questionnaire-style dialogues as training inputs with explicit character personality labels, offer strong controllability but are highly dependent on data quality and label accuracy. In contrast, self-supervised methods like Ditto \citep{chen_large_2023} do not use any personality labels. Instead, they construct WIKIROLE, a dataset containing 4,000 characters, through self-alignment, activating the model's internalized role knowledge via character self-questioning. This method can obtain highly diverse character styles without external intervention and exhibits strong generalization.

Notably, these two types of methods differ in terms of personality consistency: supervised methods are more suitable for fine-grained simulation of specific character settings, while self-supervised methods are better suited for generating open-ended, creative new characters.

\subsection{Character Memory Mechanisms}
The consistent performance of characters in long-term dialogues largely depends on whether they possess effective memory mechanisms. Human interaction relies on memory of past events, others' identities, and interaction contexts. If RPLAs cannot maintain continuity in character experiences, knowledge, and relationships, their behaviors are prone to imbalance or character breakdown.

To address this, Xu et al. \citeyearpar{xu_character_2024} proposed the CHARMAP mechanism, which explicitly represents character memory as multiple text chunks and dynamically selects memory content relevant to the current context during generation, adding it to the model input as additional prompts. This mechanism significantly improved behavioral consistency in the task of generating character decisions in novels and effectively reduced issues such as character forgetting and context confusion.

Building on the idea of CHARMAP, a generalized MAP (memory-augmented prompt) structure has emerged in recent years, dividing the model into two parts: a main generator responsible for language generation and a memory retriever responsible for retrieving relevant segments from external memory. This structure emphasizes explicit retrieval and causal embedding: memory is no longer a vaguely latent state in model weights but a module with semantic meaning that can be observed and manipulated. This structure also provides mechanical support for memory sharing or updating among multiple agents in the future.

Notably, in multi-character systems (e.g., game dialogues, multi-character scripts), balancing the independence of individual memory and the consistency of shared memory is a major challenge. Excessively shared memory may lead to character personality convergence, while excessively independent memory hinders the construction of a unified worldview and scenario logic. The Ditto system \citep{chen_large_2023} partially addressed this issue by maintaining independent character folders for each character during training while constructing a world consensus library through cross-dialogue data. This suggests that future RPLA systems may need to draw on the concept of meta-memory networks, supporting the collaborative evolution of character private memory, group shared memory, and environmental background memory.

\subsection{Character Behavior and Decision-Making Modeling}
Language generation is only the first step in role simulation. Whether a character possesses behavioral decision-making capabilities consistent with their personality is the core criterion for evaluating its intelligence and anthropomorphism. To this end, Xu et al. \citeyearpar{xu_character_2024} constructed the LIFECHOICE dataset, collecting decision-making content of novel characters in key scenarios along with their personality setting labels, and proposed the task: Given a scenario and character settings, can the model select a behavioral path consistent with the character's personality?

Experiments showed that existing language models (e.g., ChatGPT) generate highly inconsistent behavioral decisions in the absence of explicit personality prompts. However, with the introduction of CHARMAP memory, the model's consistency scores significantly improved, indicating that character behavior needs to be constrained by the causal chain of personality-motivation-situation.

Most existing RPLA systems rely on LLMs to simulate character images through dialogue generation, with character behavioral logic often implicitly integrated into the dialogue. However, from a cognitive perspective, there are mechanistic differences between linguistic expression and behavioral decision-making, and dialogue alone is insufficient to support complex, dynamic plot evolution. In response, Wu et al. \citeyearpar{wu_role-play_2024} proposed deep integration of plot construction mechanisms with language generation, introducing narrative chains and plot triggers to enable the model to possess behavior guidance and decision-making capabilities before dialogue generation, thereby realizing behavioral pre-decision based on interaction context. This design significantly enhanced the controllability and coherence of plot control and has initially demonstrated effectiveness in interactive plot generation tasks, expanding the capability boundaries of RPLA systems from language modeling to plot regulation.

From the perspective of existing research, systems with strong behavioral controllability (e.g., CHARMAP) often rely on detailed settings and explicit instructions, making them suitable for high-precision role simulation scenarios such as interactive novels and virtual theaters; while systems with strong generalization (e.g., Ditto) are more suitable for mass character generation and unlabeled open domains.

A comparison of representative systems in terms of behavioral modeling dimensions is shown in Table\ref{table1}.

\begin{table}[ht]      
\centering             
\caption{Comparison of Modeling Dimensions Across Different Systems}  
\label{table1}    
\begin{tabular}{p{1.5cm} p{2.5cm} p{2cm} p{1.7cm} p{1.8cm} p{2cm}}
\toprule               
{\bf System} & {\bf Behavior control mechanism} & {\bf Memory structure} & {\bf Personality consistency score} & {\bf generalization ability} & {\bf Applicable scenarios} \\
\midrule               
{\bf CHARMAP} & Explicit memory + situational prompts & External snippet structure & high & Medium & Novel character decision-making \\
\addlinespace
{\bf Ditto} & Self-supervised style induction & Self-built character folders & Medium to high & high & Generalized character generation \\
\addlinespace
{\bf CoSER} & Motivation-driven + given circumstances & Narrative nested worldview & high & Medium & Multi-character scripts \\
\bottomrule            
\end{tabular}
\end{table}

The capability bottleneck of RPLAs is shifting from language generation to the complex coordination of personality modeling, behavioral reasoning, and memory consistency. From a psychological perspective, a character is not merely a product of linguistic style but a comprehensive embodiment of values, emotional regulation, and goal-directed behavior. Therefore, key breakthroughs in future RPLAs will involve the introduction of emotional regulation mechanisms and conflict response simulation to achieve more dynamic personality expression. At the same time, constructing hierarchical memory architectures to ensure continuity in long-term interactions among multiple characters, and integrating behavioral planning models with multimodal generation technologies, will also become important foundations and development directions for building complete virtual character systems with linguistic, behavioral, and perceptual capabilities.

\section{Construction and Annotation of Role-Specific Data}
The modeling effect of RPLAs largely depends on the quality and structured nature of data. Unlike general text corpora required by ordinary language models, RPLA training demands high-quality role-specific corpora with personality, situational, and behavioral consistency, which typically require extracting structured role information from multimodal, multi-level raw materials. This section will elaborate on four aspects: data sources, construction processes, structural design, and open issues.

\subsection{Data Sources: Anthropomorphic Resources from Novels to Scripts}
Common corpus sources in existing RPLA research can be divided into the following categories:
\begin{enumerate}
    \item Literary and script data: Including novels, scripts, and dialogue comics. Such texts usually have clear character settings and complete plots, making them ideal materials for personality modeling. Representative projects include Xu et al. \citeyearpar{xu_character_2024} constructing the LIFECHOICE decision-making dataset based on novel data, and CoSER \citep{wang_coser_2025} extracting multi-character, multi-plot data structures from massive Chinese online novels.

    \item TV drama and film script data: TV drama and film scripts often feature continuous plot development, clear character arcs, and rich dialogue content, making them high-quality corpora for constructing dynamic, multi-faceted character personalities. For example, some character dialogue systems use subtitles or scripts from dramas such as "Legend of Zhen Huan" and "Bright Sword" to train multi-character collaborative models, simulating long-term interactions and complex relationships among characters.
    
    \item Anime and virtual character corpora: Character texts based on ACG culture (Anime, Comics, Games) have highly stylized characteristics. For instance, ChatHaruhi \citep{li_chatharuhi_2023} focuses on linguistic imitation of the anime character "Haruhi Suzumiya," with its constructed data derived from original novels, anime scripts, and user role simulation records.
    
    \item Historical and biographical resources: Suitable for building highly credible personality agents. For example, the character profile fine-tuning module in CharacterGLM \citep{zhou_characterglm_2024} uses public historical figure data to generate personalized instructions, training roles with specific knowledge and values.
\end{enumerate}
A common feature of these corpus sources is that the content is character-centric, the semantics have clear contexts, and there are specific behavioral motivations or relationship structures. Compared to pure web-based question-answering or dialogue data, such corpora are more suitable for role-level modeling.

\subsection{Construction Process: Integration of Automatic Extraction and Manual Verification}
\begin{figure}[htbp]
    \centering
    \includegraphics[width=\linewidth]{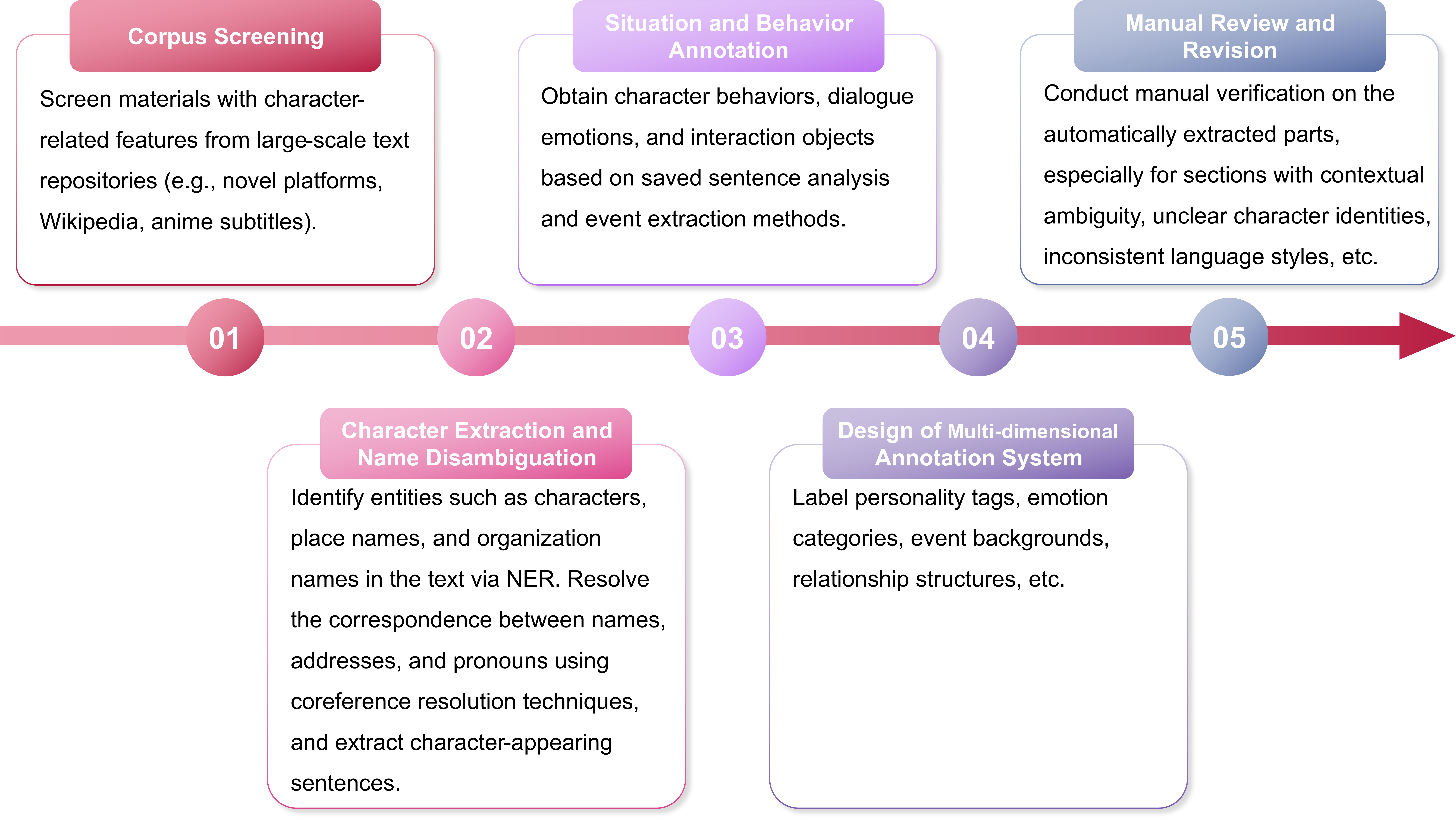}
    \caption{Construction Process of High-Quality Role-Specific Corpora}
    \label{figure1}
\end{figure}

The construction of high-quality role-specific corpora typically involves the following steps, as shown in Figure\ref{figure1}. Firstly, corpus materials with character characteristics are selected from large text libraries (e.g., novel websites, Wikipedia, and anime subtitles). Secondly, named entity recognition (NER) technology is used to identify entities such as character names, place names, and organization names in the text, and coreference resolution is employed to address correspondences between names, titles, and pronouns, thereby extracting the character's utterances. Subsequently, based on dependency parsing and event extraction methods, annotations are made on the character's behaviors, dialogue emotions, and interaction objects. On this basis, a multi-dimensional annotation system is designed, including personality labels (e.g., MBTI), emotional categories, event backgrounds, and relationship structures among characters. Finally, manual review and correction are performed on the automatically extracted and annotated content, with special attention to issues such as contextual ambiguity, unclear character identities, and inconsistent linguistic styles, to ensure the final corpus is of high quality and accuracy.

For example, when extracting the character language of Haruhi Suzumiya from original novels and anime subtitles, ChatHaruhi \citep{li_chatharuhi_2023} used automatic syntactic analysis tools to initially extract sentences, followed by manual annotation by volunteers on key dimensions such as whether the character's unique tone is maintained, whether it fits the script context, and whether others' language is mixed in. Ultimately, a training set with strong style unity and semantic consistency was constructed for training Japanese and Chinese-English bilingual RPLAs.

CoSER \citep{wang_coser_2025} proposed a data construction process of "structured extraction + narrative modeling," systematically annotating the character-motivation-behavior-consequence quadruple for each event in the novel and establishing a cross-novel character relationship graph, providing data support for complex character behavior simulation and consistency modeling.

\subsection{Data Dimension Design: Multi-Level Structure from Personality to Events}
A high-quality RPLA corpus should not only cover a sufficient number of characters and scenarios but also meet the needs of multi-level modeling in terms of data structure. Common strategies for creating character data dimensions are shown in Table\ref{table2}.

\begin{table}[ht]      
\centering             
\caption{Role data dimensions and their creation strategies}  
\label{table2}    
\begin{tabular}{p{3.7cm}p{4.3cm}p{4.5cm}}
\toprule               
{\bf Dimension} & {\bf Content} & {\bf Typical annotation methods} \\
\midrule               
{\bf Personality Trait Labels} & e.g., Big Five, MBTI & Human assessment or model prediction \\
\addlinespace
{\bf Linguistic Style} & Vocabulary us, speech rate, sentence structure, etc. & Statistical features + style clustering \\
\addlinespace
{\bf Relationship Structure} & Interaction graph with other characters & Manual annotation or graph structure construction \\
\addlinespace
{\bf Situational Context} & Event background, emotional trends & Event extraction + emotion classification \\
\addlinespace
{\bf Behavioral Records} & Decision paths, action plans & Action extraction + intent recognition \\
\bottomrule            
\end{tabular}
\end{table}

The CoSER \citep{wang_coser_2025} dataset innovatively proposed a narrative modeling approach of "given-circumstance + behavior generation," where each character generates behaviors within a structured event node, helping the model learn the causal structure of "character behavior driven by motivation and situation."

\subsection{Trade-Offs Between Open and Private Corpora}
Although the technology for constructing role-specific data is relatively mature, challenges remain in terms of openness, authorization, and reusability. Firstly, copyright issues constitute a significant constraint. Many high-quality corpora are derived from novels, scripts, and anime, but the ownership of these contents is often unclear or not publicly licensed, making it difficult to open-source the datasets. For example, the data of ChatHaruhi \citep{li_chatharuhi_2023} can only be used for research purposes, not for commercial use. Secondly, the issue of style unity is prominent. Corpora from different character sources vary greatly in style, and cross-corpus training may lead to blurred character personalities. Although Ditto \citep{chen_large_2023} attempted to mitigate this problem through self-supervised style modeling, it still cannot fully resolve the issue of character style drift. Finally, the scarcity of open resources is a practical problem. Currently, there are relatively few high-quality, structured role datasets that are truly open and reproducible. While datasets such as InCharacter \citep{wang_incharacter_2024} and CharacterEval \citep{tu_charactereval_2024} have open-sourced evaluation scripts and a small number of samples, most open datasets have limitations in terms of scale or quality.

Therefore, future research needs to make improvements in two important directions. Firstly, there is a need to promote the openness of role datasets and formulate clear norms for role dataset construction to ensure data reusability and openness. Secondly, the role of multimodal role data (e.g., speech, video) in RPLA training should be explored to enrich the expressive dimensions of characters.

In summary, role data construction is not only an input source for model training but also a fundamental guarantee for RPLAs to possess human-like qualities. From the selection of raw materials, structured modeling to the embedding of behavioral motivations, current research has formed various technical routes. Future directions should focus on 1) constructing high-dimensional, structured, and controllable character knowledge bases; 2) promoting the construction of an open data ecosystem to address barriers to authorization and reuse; 3) introducing generative data augmentation mechanisms to improve the data diversity and situational generalization capabilities of role modeling.

\section{Evaluation Methods and Metrics}
A core challenge for RPLAs lies not only in the fluency and context relevance of generated language but also in maintaining the consistency of character personality traits over multi-turn dialogues, ensuring the accuracy of background knowledge, and the rationality of behavioral decisions. Such challenges inherently involve the collaborative capabilities of cognitive modeling, semantic control, and situational awareness. Therefore, establishing a scientific, systematic, and scalable evaluation system is not only an important means of measuring model performance but also the foundation for achieving application credibility and user experience optimization. Currently, the evaluation system of RPLAs can be divided into two dimensions: evaluation metrics and evaluation methods.

\subsection{Construction of Evaluation Metrics and Benchmark Systems}
In recent years, several research teams have strived to develop evaluation benchmarks for RPLAs. These benchmark systems not only cover the testing of role knowledge but also involve the multi-dimensional evaluation of role-playing capabilities.

In terms of role knowledge evaluation, RoleEval proposed by Shen et al. \citeyearpar{shen_roleeval_2024} represents an important milestone, aiming to systematically assess LLMs' abilities to memorize, utilize, and reason about role knowledge. RoleEval consists of RoleEval-Global and RoleEval-Chinese, covering 300 characters from different domains (e.g., celebrities, anime, movies) and comprehensively examining the character's personal information, relationships, abilities, and experiences through 6,000 Chinese-English bilingual multiple-choice questions. This benchmark ensures the diversity and challenge of questions through a combination of automatic and manual verification, revealing significant differences in role knowledge among different language models.

In addition to the evaluation of role knowledge, multi-dimensional evaluation of role-playing capabilities has gradually gained attention. CharacterEval proposed by Tu et al. \citeyearpar{tu_charactereval_2024} is a benchmark for Chinese role-playing conversational agents, containing 1,785 multi-turn role-playing dialogues covering 77 characters from Chinese novels and scripts. CharacterEval conducts a comprehensive evaluation through 13 metrics across four dimensions: dialogue ability, role consistency, role-playing appeal, and personality recall test, and develops a human-annotated role-playing reward model CharacterRM to more accurately measure the quality of role-playing dialogues.

Furthermore, the RoleLLM framework \citep{wang_rolellm_2024} has further expanded the evaluation of role-playing capabilities. Through four stages—character profile construction, context-based instruction generation, role prompting, and role-conditioned instruction fine-tuning—the framework systematically evaluates and enhances LLMs' role-playing capabilities. RoleLLM constructs the RoleBench dataset containing 168,093 samples for the systematic evaluation of role-playing. Through fine-tuning on RoleBench, RoleLLM significantly improves the role-playing capabilities of open-source models, even reaching levels comparable to GPT-4 in certain aspects.

Role consistency is an important aspect of role-playing. InCharacter \citep{wang_incharacter_2024} evaluates the role consistency of role-playing agents through psychological measurement scales. This study assesses 32 different characters using 14 widely used psychological measurement scales, verifying the effectiveness of InCharacter in measuring the personality of RPLAs. Experimental results show that state-of-the-art RPLAs can highly accurately reproduce the personality of target characters, with an accuracy rate of up to 80.7\%.

In terms of the interactivity of role-playing, SHARP proposed by Kong et al. \citeyearpar{kong_sharp_2024} is a benchmark constructed based on interactive hallucination of stance transfer in RPLAs. This study extracts relationships from common-sense knowledge graphs and uses the inherent hallucination properties of RPLAs to simulate cross-character interactions, constructing a general and effective benchmark for automatically measuring role fidelity. Experiments verify the effectiveness and stability of this paradigm and further explore the factors influencing these metrics.

In addition, TimeChara proposed by Ahn et al. \citeyearpar{ahn_timechara_2024} emphasizes the importance of point-in-time role-playing. This benchmark generates 10,895 instances through an automated process to evaluate the point-in-time character hallucination problem of role-playing LLMs. The study finds that current state-of-the-art LLMs (e.g., GPT-4) have significant issues with point-in-time character hallucination. To address this challenge, the study proposes the Narrative-Experts method, which effectively reduces point-in-time character hallucination by decomposing reasoning steps and utilizing narrative experts.

With the rapid development of role-playing dialogue systems, user-centric evaluation methods have gradually gained attention. RMTBench  \citep{xiang_rmtbench_2025} is a novel user-centric role-playing benchmark containing 80 diverse characters and over 8,000 rounds of dialogue data. Unlike previous role-centric evaluation methods, RMTBench constructs dialogues based on user-centric scenarios, exploring the impact of shifting the dialogue focus from characters to users on model performance. In addition, RMTBench implements a multi-dimensional automatic evaluation system and conducts extensive analyses and experiments. By emphasizing user centricity and multi-dimensional scenarios, RMTBench provides an important supplement for establishing role-playing benchmarks more in line with practical applications.

In role-playing, a character's values play a fundamental role in their behaviors and attitudes. Therefore, the consistency of values is crucial for enhancing the realism of interactions and enriching user experience. RVBench proposed by Wang et al. \citeyearpar{wang_rvbench_2025} effectively fills this gap. This benchmark is an evaluation benchmark for character values, containing 25 characters as benchmark data, and proposes evaluation methods including value scoring and value ranking by drawing on human psychological test methods. The value scoring method tests the character's value orientation through the Revised Portrait Values Questionnaire (PVQ-RR), providing direct and quantitative character comparisons. The value ranking method evaluates whether the agent's behaviors in dilemma scenarios are consistent with the hierarchical structure of their values. Tests on various open-source and closed-source LLMs show that GLM-4's values are closest to those of the characters in the benchmark data. However, compared to preset characters, LLMs' role-playing capabilities still have certain gaps, including in the consistency, stability, and flexibility of values. These findings urgently require further research to improve LLMs' role-playing capabilities from the perspective of value alignment.

In summary, significant progress has been made in the construction of evaluation metrics and benchmark systems in the field of role-playing. From the systematic evaluation of role knowledge to the multi-dimensional measurement of role-playing capabilities, and the in-depth research on role consistency and interactivity issues, these benchmark systems provide important evaluation tools for the application of LLMs in the field of role-playing. However, current research also reveals the challenges faced by existing models in terms of character hallucination, point-in-time role-playing, and value alignment, which provides new directions for future research.

\subsection{Evaluation Methods and Reward Mechanism Modeling}
Although automatic evaluation offers scalability and efficiency advantages, judging aspects such as character realism, linguistic emotion, and personality charm still relies on human subjective experience. In recent years, "semi-automatic subjective evaluation" combining human scoring and reward model construction has become the mainstream approach.

Human evaluation is currently the most intuitive and interpretable method for RPLA assessment, typically relying on expert annotators to score model-generated results based on preset scales. This method offers high flexibility, enabling the capture of subtle performances and inconsistent details in model outputs, but it also suffers from high costs, strong subjectivity, and poor repeatability. Meanwhile, reinforcement learning from human feedback (RLHF) has gradually become an important supplementary tool for evaluating RPLAs. By collecting human subjective preference data on generated samples, a reward model is trained to simulate human scoring behavior and guide model optimization. Among them, CharacterRM proposed by Tu et al. \citeyearpar{tu_charactereval_2024} is the first reward model specifically designed for RPLAs, integrating dialogue ability, role consistency, role-playing appeal, and personality recall test as reward signals. This model outperforms GPT-4 in terms of correlation with human judgments. However, such methods still have certain limitations. The most significant issue is that reward models generally lack generalization capabilities, often requiring separate training for each type of character or evaluation target, resulting in high evaluation costs and poor transferability. Therefore, improving the versatility and cross-character adaptability of reward models remains a key challenge in current research.

With the continuous improvement of the capabilities of large language models (e.g., GPT-4), using them as automatic scorers to evaluate the generated content of RPLAs has gradually become one of the mainstream methods. This approach typically assigns the model the role of a character evaluator, guiding it to score or rank candidate responses in terms of role consistency, linguistic fluency, and knowledge accuracy through natural language prompts, thereby achieving an automated, low-cost, and scalable evaluation process.

However, existing research has questioned the effectiveness and fairness of such methods. For example, Yuan et al. \citeyearpar{yuan_evaluating_2024} proposed that GPT-like scorers tend to favor linguistic expression quality during scoring, ignoring or downplaying the fidelity to character settings. Specifically, outputs that are more natural and fluent in language but violate character settings may be assigned high scores, leading to distorted evaluation results. In addition, LLMs themselves have limitations in character understanding and behavioral logic grasp, and their scoring judgments lack clear explanation mechanisms and verifiability, resulting in widespread concerns about the credibility of evaluation results.

Therefore, although LLM-based automatic scoring brings convenience and automation advantages to RPLA evaluation, its evaluation bias also suggests that we need to further optimize the design and application of this method in terms of prompt design, alignment mechanisms, and multi-dimensional scoring constraints. Possible future development directions include multi-model cross-validation, the introduction of multi-round scoring frameworks, and enhancing scoring credibility through expert annotation integration.

The evaluation system of RPLAs is gradually shifting from simply focusing on language quality to assessing personality consistency and behavioral rationality. This not only requires quantitative scoring mechanisms but also relies on subjective aesthetic judgments. In the future, a three-dimensional evaluation system integrating "personality alignment—behavior detection—situational integration" should be developed, introducing more refined psychological evaluation frameworks (e.g., dynamic personality tests), more multi-dimensional composite scoring systems, and long-term character behavior log analysis tools that can be continuously tracked, thereby achieving a comprehensive, long-term, and dynamic evaluation of RPLAs.

\section{Future Outlook and Research Trends}
With the continuous development of the field of RPLAs driven by LLMs, future research is gradually moving from static personality modeling toward more dynamic and complex personality evolution modeling. This trend not only makes RPLAs more vibrant and realistic in performance but also provides new possibilities for the deep integration of artificial intelligence with human emotions and cognition.

Traditional RPLAs mostly adopt fixed personality templates, maintaining character stability and consistency through preset personality traits and behavioral rules. However, real personalities are not static but constantly evolve over time and in different environments. Future research will introduce meta-learning and emotion modeling to enable role agents to self-adjust and evolve during interactions. Meta-learning provides agents with the ability to "learn how to learn," allowing them to quickly adapt and optimize their personality performance even when encountering new scenarios or user feedback. Combined with emotion modeling, agents can make adjustments not only at the knowledge level but also achieve delicate changes in emotional experience and expression, making personalities more layered and in-depth. Such evolutionary personalities help create more vivid and engaging characters, enhancing user immersion and emotional connection.

Future RPLA research will also focus on the development of multi-agent systems. In a complex story or virtual world, the behaviors of different characters interact, collaborate, or conflict, forming a multi-level, multi-threaded narrative structure. For example, Wu et al. \citeyearpar{wu_role-play_2024} have attempted to allow multiple AI characters to participate in story generation together, resulting in richer and more unpredictable plot developments. The collaboration and confrontation mechanisms of multi-agents not only improve the diversity and authenticity of stories but also place higher demands on character consistency management and emotional interaction. Future research will explore how to achieve dynamic role relationship management, information sharing, and conflict mediation among agents, ensuring coherent stories and distinctive character images.

Pure text interaction has gradually been unable to meet users' demands for immersive experiences. The integration of multimodal information, such as voice timbre, facial expressions, and body gestures, will greatly enhance the expressiveness and sense of presence of RPLAs. Future research will utilize technologies such as computer vision, speech recognition, and motion capture to enable agents to express emotions and intentions through richer channels. For example, when an agent is emotionally excited, its facial expressions will be more vivid, its voice intonation more engaging, and its body movements smooth and natural. This multimodal integration will break through the limitations of traditional dialogue, making virtual characters closer to real human interaction experiences, and finding wide applications in fields such as virtual anchors, digital humans, game NPCs, and psychological counseling.

Personality formation and behavioral decision-making are complex neurocognitive processes. Future RPLA design will more deeply integrate research findings from cognitive neuroscience to simulate real human decision-making mechanisms and emotional responses. By drawing on neuroscientific understanding of brain functional areas, neural circuits, and their dynamic interactions, agents can achieve behavioral patterns more in line with human psychological mechanisms. For example, emotion regulation models based on neural circuits can enable characters to produce more reasonable emotional fluctuations and behavioral choices in different scenarios; using cognitive load and attention mechanism simulations, agents can more naturally handle multi-task information and priority judgments. This direction not only promotes the scientization and refinement of AI personality models but also fosters interdisciplinary integration between artificial intelligence and human psychology, psychiatry, laying the foundation for building agents with human-like perception and decision-making characteristics.

In summary, the future development trends of RPLAs will present a multi-dimensional integration pattern of dynamic personality evolution, complex multi-agent interaction, multimodal immersive experience, and integration with cognitive neuroscience. Only by continuously expanding these cutting-edge fields can virtual characters approach real life, meeting people's higher expectations for intelligent companionship, emotional communication, and immersive experiences.

\section{Conclusion}
This paper systematically reviews the research progress and challenges in the field of RPLAs driven by LLMs. From personality modeling technologies and interaction mechanisms to multimodal integration, various technologies have continuously promoted significant improvements in the realism and interaction depth of virtual characters. However, existing research still faces numerous bottlenecks, including the maintenance of personality consistency, conflict management in multi-character collaboration, the diversity and robustness of user inputs, and the complexity of ethical and legal risks. These issues limit the promotion and popularization of RPLAs in broader and more complex application scenarios.

Nevertheless, RPLA technology has demonstrated strong technical potential and far-reaching social impacts. From a technical perspective, RPLAs not only promote the integrated development of cutting-edge artificial intelligence technologies such as NLP, multimodal fusion, and reinforcement learning but also drive breakthroughs in agent personalization and emotional interaction, greatly enhancing the interaction experience and adaptability of virtual characters. From a social perspective, RPLAs have shown broad application prospects in fields such as game entertainment, educational companionship, psychological counseling, and digital humans, providing people with new ways of communication and companionship and influencing all aspects of digital life.

In the future, building multi-lingual, multi-cultural, and multi-perspective character systems will become an important direction for the development of RPLAs. In the context of globalization, personality expression, values, and interaction habits vary across different cultural and linguistic environments. How to enable agents to cross cultural boundaries and achieve true personalization and diversification is an urgent issue to be addressed. At the same time, emphasizing multi-perspective design helps avoid biases from a single cultural perspective and improves the inclusiveness and fairness of the system.

In conclusion, as a model of interdisciplinary integration between artificial intelligence and the humanities and social sciences, RPLAs face both challenges and opportunities. Only through interdisciplinary collaboration, technological innovation, and equal emphasis on social responsibility can RPLAs move toward a smarter, more realistic, and more inclusive future.

\bibliographystyle{plainnat}
\bibliography{references}

@inproceedings{zhou_characterglm_2024,
	address = {Miami, Florida, US},
	title = {{CharacterGLM}: {Customizing} {Social} {Characters} with {Large} {Language} {Models}},
	shorttitle = {{CharacterGLM}},
	url = {https://aclanthology.org/2024.emnlp-industry.107/},
	doi = {10.18653/v1/2024.emnlp-industry.107},
	urldate = {2025-06-11},
	booktitle = {Proceedings of the 2024 {Conference} on {Empirical} {Methods} in {Natural} {Language} {Processing}: {Industry} {Track}},
	publisher = {Association for Computational Linguistics},
	author = {Zhou, Jinfeng and Chen, Zhuang and Wan, Dazhen and Wen, Bosi and Song, Yi and Yu, Jifan and Huang, Yongkang and Ke, Pei and Bi, Guanqun and Peng, Libiao and Yang, JiaMing and Xiao, Xiyao and Sabour, Sahand and Zhang, Xiaohan and Hou, Wenjing and Zhang, Yijia and Dong, Yuxiao and Wang, Hongning and Tang, Jie and Huang, Minlie},
	editor = {Dernoncourt, Franck and Preoţiuc-Pietro, Daniel and Shimorina, Anastasia},
	month = nov,
	year = {2024},
	pages = {1457--1476},
}

@inproceedings{ran_capturing_2024,
	address = {Miami, Florida, USA},
	title = {Capturing {Minds}, {Not} {Just} {Words}: {Enhancing} {Role}-{Playing} {Language} {Models} with {Personality}-{Indicative} {Data}},
	shorttitle = {Capturing {Minds}, {Not} {Just} {Words}},
	url = {https://aclanthology.org/2024.findings-emnlp.853},
	doi = {10.18653/v1/2024.findings-emnlp.853},
	language = {en},
	urldate = {2025-06-11},
	booktitle = {Findings of the {Association} for {Computational} {Linguistics}: {EMNLP} 2024},
	publisher = {Association for Computational Linguistics},
	author = {Ran, Yiting and Wang, Xintao and Xu, Rui and Yuan, Xinfeng and Liang, Jiaqing and Xiao, Yanghua and Yang, Deqing},
	year = {2024},
	pages = {14566--14576},
}

@misc{xu_character_2024,
	title = {Character is {Destiny}: {Can} {Role}-{Playing} {Language} {Agents} {Make} {Persona}-{Driven} {Decisions}?},
	shorttitle = {Character is {Destiny}},
	url = {http://arxiv.org/abs/2404.12138},
	doi = {10.48550/arXiv.2404.12138},
	language = {en-US},
	urldate = {2025-03-05},
	publisher = {arXiv},
	author = {Xu, Rui and Wang, Xintao and Chen, Jiangjie and Yuan, Siyu and Yuan, Xinfeng and Liang, Jiaqing and Chen, Zulong and Dong, Xiaoqing and Xiao, Yanghua},
	month = nov,
	year = {2024},
	note = {arXiv:2404.12138 [cs]},
}

@misc{curry_characterai_2025,
  author       = {Curry, David},
  title        = {Character.AI Revenue and Usage Statistics (2025)},
  howpublished = {\url{https://www.businessofapps.com/data/character-ai-statistics/}},
  year         = {2025},
  note         = {Business of Apps; Accessed: 2025-05-14}
}

@inproceedings{wang_incharacter_2024,
	address = {Bangkok, Thailand},
	title = {{InCharacter}: {Evaluating} {Personality} {Fidelity} in {Role}-{Playing} {Agents} through {Psychological} {Interviews}},
	shorttitle = {{InCharacter}},
	url = {https://aclanthology.org/2024.acl-long.102/},
	doi = {10.18653/v1/2024.acl-long.102},
	urldate = {2025-06-11},
	booktitle = {Proceedings of the 62nd {Annual} {Meeting} of the {Association} for {Computational} {Linguistics} ({Volume} 1: {Long} {Papers})},
	publisher = {Association for Computational Linguistics},
	author = {Wang, Xintao and Xiao, Yunze and Huang, Jen-tse and Yuan, Siyu and Xu, Rui and Guo, Haoran and Tu, Quan and Fei, Yaying and Leng, Ziang and Wang, Wei and Chen, Jiangjie and Li, Cheng and Xiao, Yanghua},
	editor = {Ku, Lun-Wei and Martins, Andre and Srikumar, Vivek},
	month = aug,
	year = {2024},
	pages = {1840--1873},
}

@inproceedings{tu_charactereval_2024,
	address = {Bangkok, Thailand},
	title = {{CharacterEval}: {A} {Chinese} {Benchmark} for {Role}-{Playing} {Conversational} {Agent} {Evaluation}},
	shorttitle = {{CharacterEval}},
	url = {https://aclanthology.org/2024.acl-long.638/},
	doi = {10.18653/v1/2024.acl-long.638},
	urldate = {2025-06-11},
	booktitle = {Proceedings of the 62nd {Annual} {Meeting} of the {Association} for {Computational} {Linguistics} ({Volume} 1: {Long} {Papers})},
	publisher = {Association for Computational Linguistics},
	author = {Tu, Quan and Fan, Shilong and Tian, Zihang and Shen, Tianhao and Shang, Shuo and Gao, Xin and Yan, Rui},
	editor = {Ku, Lun-Wei and Martins, Andre and Srikumar, Vivek},
	month = aug,
	year = {2024},
	pages = {11836--11850},
}

@inproceedings{si_telling_2021,
	address = {Singapore and Online},
	title = {Telling {Stories} through {Multi}-{User} {Dialogue} by {Modeling} {Character} {Relations}},
	url = {https://aclanthology.org/2021.sigdial-1.30/},
	doi = {10.18653/v1/2021.sigdial-1.30},
	language = {en-US},
	urldate = {2025-06-11},
	booktitle = {Proceedings of the 22nd {Annual} {Meeting} of the {Special} {Interest} {Group} on {Discourse} and {Dialogue}},
	publisher = {Association for Computational Linguistics},
	author = {Si, Wai Man and Ammanabrolu, Prithviraj and Riedl, Mark},
	editor = {Li, Haizhou and Levow, Gina-Anne and Yu, Zhou and Gupta, Chitralekha and Sisman, Berrak and Cai, Siqi and Vandyke, David and Dethlefs, Nina and Wu, Yan and Li, Junyi Jessy},
	month = jul,
	year = {2021},
	pages = {269--275},
}

@misc{vaswani_attention_2023,
	title = {Attention {Is} {All} {You} {Need}},
	url = {http://arxiv.org/abs/1706.03762},
	doi = {10.48550/arXiv.1706.03762},
	language = {en-US},
	urldate = {2025-03-04},
	publisher = {arXiv},
	author = {Vaswani, Ashish and Shazeer, Noam and Parmar, Niki and Uszkoreit, Jakob and Jones, Llion and Gomez, Aidan N. and Kaiser, Lukasz and Polosukhin, Illia},
	month = aug,
	year = {2023},
	note = {arXiv:1706.03762 [cs]},
}

@inproceedings{shao_character-llm_2023,
	address = {Singapore},
	title = {Character-{LLM}: {A} {Trainable} {Agent} for {Role}-{Playing}},
	shorttitle = {Character-{LLM}},
	url = {https://aclanthology.org/2023.emnlp-main.814},
	doi = {10.18653/v1/2023.emnlp-main.814},
	language = {en},
	urldate = {2025-06-11},
	booktitle = {Proceedings of the 2023 {Conference} on {Empirical} {Methods} in {Natural} {Language} {Processing}},
	publisher = {Association for Computational Linguistics},
	author = {Shao, Yunfan and Li, Linyang and Dai, Junqi and Qiu, Xipeng},
	year = {2023},
	pages = {13153--13187},
}

@misc{li_chatharuhi_2023,
	title = {{ChatHaruhi}: {Reviving} {Anime} {Character} in {Reality} via {Large} {Language} {Model}},
	shorttitle = {{ChatHaruhi}},
	url = {http://arxiv.org/abs/2308.09597},
	doi = {10.48550/arXiv.2308.09597},
	urldate = {2025-05-22},
	publisher = {arXiv},
	author = {Li, Cheng and Leng, Ziang and Yan, Chenxi and Shen, Junyi and Wang, Hao and MI, Weishi and Fei, Yaying and Feng, Xiaoyang and Yan, Song and Wang, HaoSheng and Zhan, Linkang and Jia, Yaokai and Wu, Pingyu and Sun, Haozhen},
	month = aug,
	year = {2023},
	note = {arXiv:2308.09597 [cs]},
}

@inproceedings{chen_large_2023,
	address = {Singapore},
	title = {Large {Language} {Models} {Meet} {Harry} {Potter}: {A} {Dataset} for {Aligning} {Dialogue} {Agents} with {Characters}},
	shorttitle = {Large {Language} {Models} {Meet} {Harry} {Potter}},
	url = {https://aclanthology.org/2023.findings-emnlp.570/},
	doi = {10.18653/v1/2023.findings-emnlp.570},
	language = {en-US},
	urldate = {2025-05-23},
	booktitle = {Findings of the {Association} for {Computational} {Linguistics}: {EMNLP} 2023},
	publisher = {Association for Computational Linguistics},
	author = {Chen, Nuo and Wang, Yan and Jiang, Haiyun and Cai, Deng and Li, Yuhan and Chen, Ziyang and Wang, Longyue and Li, Jia},
	editor = {Bouamor, Houda and Pino, Juan and Bali, Kalika},
	month = dec,
	year = {2023},
	pages = {8506--8520},
}

@misc{wang_coser_2025,
	title = {{CoSER}: {Coordinating} {LLM}-{Based} {Persona} {Simulation} of {Established} {Roles}},
	shorttitle = {{CoSER}},
	url = {https://arxiv.org/abs/2502.09082v2},
	language = {en},
	urldate = {2025-06-11},
	journal = {arXiv.org},
	author = {Wang, Xintao and Wang, Heng and Zhang, Yifei and Yuan, Xinfeng and Xu, Rui and Huang, Jen-tse and Yuan, Siyu and Guo, Haoran and Chen, Jiangjie and Zhou, Shuchang and Wang, Wei and Xiao, Yanghua},
	month = feb,
	year = {2025},
}

@misc{cheng_psymem_2025,
	title = {{PsyMem}: {Fine}-grained psychological alignment and {Explicit} {Memory} {Control} for {Advanced} {Role}-{Playing} {LLMs}},
	shorttitle = {{PsyMem}},
	url = {http://arxiv.org/abs/2505.12814},
	doi = {10.48550/arXiv.2505.12814},
	urldate = {2025-06-05},
	publisher = {arXiv},
	author = {Cheng, Xilong and Qin, Yunxiao and Tan, Yuting and Li, Zhengnan and Wang, Ye and Xiao, Hongjiang and Zhang, Yuan},
	month = may,
	year = {2025},
	note = {arXiv:2505.12814 [cs]},
}

@inproceedings{wu_role-play_2024,
	address = {Bangkok, Thailand},
	title = {From {Role}-{Play} to {Drama}-{Interaction}: {An} {LLM} {Solution}},
	shorttitle = {From {Role}-{Play} to {Drama}-{Interaction}},
	url = {https://aclanthology.org/2024.findings-acl.196/},
	doi = {10.18653/v1/2024.findings-acl.196},
	urldate = {2025-06-11},
	booktitle = {Findings of the {Association} for {Computational} {Linguistics}: {ACL} 2024},
	publisher = {Association for Computational Linguistics},
	author = {Wu, Weiqi and Wu, Hongqiu and Jiang, Lai and Liu, Xingyuan and Zhao, Hai and Zhang, Min},
	editor = {Ku, Lun-Wei and Martins, Andre and Srikumar, Vivek},
	month = aug,
	year = {2024},
	pages = {3271--3290},
}

@misc{shen_roleeval_2024,
	title = {{RoleEval}: {A} {Bilingual} {Role} {Evaluation} {Benchmark} for {Large} {Language} {Models}},
	shorttitle = {{RoleEval}},
	url = {http://arxiv.org/abs/2312.16132},
	doi = {10.48550/arXiv.2312.16132},
	language = {en-US},
	urldate = {2025-03-05},
	publisher = {arXiv},
	author = {Shen, Tianhao and Li, Sun and Tu, Quan and Xiong, Deyi},
	month = feb,
	year = {2024},
	note = {arXiv:2312.16132 [cs]},
}

@inproceedings{wang_rolellm_2024,
	address = {Bangkok, Thailand},
	title = {{RoleLLM}: {Benchmarking}, {Eliciting}, and {Enhancing} {Role}-{Playing} {Abilities} of {Large} {Language} {Models}},
	shorttitle = {{RoleLLM}},
	url = {https://aclanthology.org/2024.findings-acl.878/},
	doi = {10.18653/v1/2024.findings-acl.878},
	language = {en-US},
	urldate = {2025-06-11},
	booktitle = {Findings of the {Association} for {Computational} {Linguistics}: {ACL} 2024},
	publisher = {Association for Computational Linguistics},
	author = {Wang, Noah and Peng, Z.y. and Que, Haoran and Liu, Jiaheng and Zhou, Wangchunshu and Wu, Yuhan and Guo, Hongcheng and Gan, Ruitong and Ni, Zehao and Yang, Jian and Zhang, Man and Zhang, Zhaoxiang and Ouyang, Wanli and Xu, Ke and Huang, Wenhao and Fu, Jie and Peng, Junran},
	editor = {Ku, Lun-Wei and Martins, Andre and Srikumar, Vivek},
	month = aug,
	year = {2024},
	pages = {14743--14777},
}

@misc{kong_sharp_2024,
	title = {{SHARP}: {Unlocking} {Interactive} {Hallucination} via {Stance} {Transfer} in {Role}-{Playing} {Agents}},
	shorttitle = {{SHARP}},
	url = {http://arxiv.org/abs/2411.07965},
	doi = {10.48550/arXiv.2411.07965},
	language = {en-US},
	urldate = {2025-03-05},
	publisher = {arXiv},
	author = {Kong, Chuyi and Luo, Ziyang and Lin, Hongzhan and Fan, Zhiyuan and Fan, Yaxin and Sun, Yuxi and Ma, Jing},
	month = dec,
	year = {2024},
	note = {arXiv:2411.07965 [cs]},
}

@inproceedings{ahn_timechara_2024,
	address = {Bangkok, Thailand},
	title = {{TimeChara}: {Evaluating} {Point}-in-{Time} {Character} {Hallucination} of {Role}-{Playing} {Large} {Language} {Models}},
	shorttitle = {{TimeChara}},
	url = {https://aclanthology.org/2024.findings-acl.197/},
	doi = {10.18653/v1/2024.findings-acl.197},
	urldate = {2025-06-11},
	booktitle = {Findings of the {Association} for {Computational} {Linguistics}: {ACL} 2024},
	publisher = {Association for Computational Linguistics},
	author = {Ahn, Jaewoo and Lee, Taehyun and Lim, Junyoung and Kim, Jin-Hwa and Yun, Sangdoo and Lee, Hwaran and Kim, Gunhee},
	editor = {Ku, Lun-Wei and Martins, Andre and Srikumar, Vivek},
	month = aug,
	year = {2024},
	pages = {3291--3325},
}

@misc{xiang_rmtbench_2025,
	title = {{RMTBench}: {Benchmarking} {LLMs} {Through} {Multi}-{Turn} {User}-{Centric} {Role}-{Playing}},
	shorttitle = {{RMTBench}},
	url = {http://arxiv.org/abs/2507.20352},
	doi = {10.48550/arXiv.2507.20352},
	language = {en-US},
	urldate = {2026-01-15},
	publisher = {arXiv},
	author = {Xiang, Hao and Tang, Tianyi and Su, Yang and Yu, Bowen and Yang, An and Huang, Fei and Zhang, Yichang and Lu, Yaojie and Lin, Hongyu and Han, Xianpei and Zhou, Jingren and Lin, Junyang and Sun, Le},
	month = oct,
	year = {2025},
	note = {arXiv:2507.20352 [cs]},
}

@article{wang_rvbench_2025,
	title = {{RVBench}: {Role} values benchmark for role-playing {LLMs}},
	volume = {5},
	issn = {29498821},
	shorttitle = {{RVBench}},
	url = {https://linkinghub.elsevier.com/retrieve/pii/S2949882125000684},
	doi = {10.1016/j.chbah.2025.100184},
	language = {en},
	urldate = {2025-08-06},
	journal = {Computers in Human Behavior: Artificial Humans},
	author = {Wang, Ye and Li, Tong and Li, Meixuan and Cheng, Ziyue and Wang, Ge and Kang, Hanyue and Deng, Yaling and Xiao, Hongjiang and Zhang, Yuan},
	month = jul,
	year = {2025},
	pages = {100184},
}

@inproceedings{yuan_evaluating_2024,
	address = {Miami, Florida, USA},
	title = {Evaluating {Character} {Understanding} of {Large} {Language} {Models} via {Character} {Profiling} from {Fictional} {Works}},
	url = {https://aclanthology.org/2024.emnlp-main.456/},
	doi = {10.18653/v1/2024.emnlp-main.456},
	urldate = {2025-06-11},
	booktitle = {Proceedings of the 2024 {Conference} on {Empirical} {Methods} in {Natural} {Language} {Processing}},
	publisher = {Association for Computational Linguistics},
	author = {Yuan, Xinfeng and Yuan, Siyu and Cui, Yuhan and Lin, Tianhe and Wang, Xintao and Xu, Rui and Chen, Jiangjie and Yang, Deqing},
	editor = {Al-Onaizan, Yaser and Bansal, Mohit and Chen, Yun-Nung},
	month = nov,
	year = {2024},
	pages = {8015--8036},
}

\end{document}